\documentclass[sigconf]{acmart}

\makeatletter
\renewcommand\@formatdoi[1]{\ignorespaces}
\makeatother

\renewcommand\footnotetextcopyrightpermission[1]{} 
\usepackage{balance}

\copyrightyear{2019}
\acmYear{2019}
\setcopyright{acmcopyright}
\acmConference[K-CAP '19]{Proceedings of the 10th International Conference on Knowledge Capture}{November 19--21, 2019}{Marina Del Rey, CA, USA}
\acmBooktitle{Proceedings of the 10th International Conference on Knowledge Capture (K-CAP '19), November 19--21, 2019, Marina Del Rey, CA, USA}
\acmPrice{15.00}
\acmDOI{10.1145/3360901.3364441}
\acmISBN{978-1-4503-7008-0/19/11}

\usepackage{array}
\usepackage{booktabs}
\usepackage{multirow}

\usepackage{fixltx2e}

\usepackage{graphicx}
\usepackage{pict2e}
\usepackage{subcaption}
\usepackage{epstopdf}
\newcommand{\fb}{\textit{FB15K-237}}
\newcommand{\wn}{\textit{WN18RR}}
\newcommand{\norm}[1]{\left\lVert#1\right\rVert}

\begin{document}

\fancyhead{}

\title[TransGCN]{TransGCN:Coupling Transformation Assumptions with Graph Convolutional Networks for Link Prediction}


\author{Ling Cai, Bo Yan, Gengchen Mai, Krzysztof Janowicz, Rui Zhu}
\affiliation{%
	\institution{STKO Lab, University of California, Santa Barbara}
}
\email{{ling.cai, boyan, gengchen_mai, jano, ruizhu}@geog.ucsb.edu}

%
\renewcommand{\shortauthors}{Cai, et al.}

%
\begin{abstract}
 Link prediction is an important and frequently studied task that contributes to an understanding of the structure of knowledge graphs (KGs) in statistical relational learning. Inspired by the success of graph convolutional networks (GCN) in modeling graph data, we propose a unified GCN framework, named TransGCN, to address this task, in which relation and entity embeddings are learned simultaneously. To handle heterogeneous relations in KGs, we introduce a novel way of representing heterogeneous neighborhood by introducing transformation assumptions on the relationship between the subject, the relation, and the object of a triple. Specifically, a relation is treated as a transformation operator transforming a head entity to a tail entity. Both translation assumption in TransE and rotation assumption in RotatE are explored in our framework. Additionally, instead of only learning entity embeddings in the convolution-based encoder while learning relation embeddings in the decoder as done by the state-of-art models, e.g., R-GCN, the TransGCN framework trains relation embeddings and entity embeddings simultaneously during the graph convolution operation, thus having fewer parameters compared with R-GCN. Experiments show that our models outperform the-state-of-arts methods on both \fb\ and \wn.
\end{abstract}

%
%


%

%

%
\maketitle

\thispagestyle{empty}

\section{Introduction}

Knowledge graphs (KGs) such as DBpedia and Freebase that encode statements about the world around us  have attracted growing attention from multiple fields, including question answering \cite{Hamilton2018EmbeddingLQ,mai2019relaxing}, knowledge inference \cite{neelakantan2015compositional}, recommendation systems \cite{Ying:2018:GCN:3219819.3219890}, and so on. By their very nature KGs are far from complete as the state of the world evolves constantly. This has motivated work on automatically predicting new statements based on known statements. Among these inference tasks link prediction has become a main focus of statistical relational learning (SRL) \cite{koller2007introduction}.

A KG encodes structural information about entities and the abundant relations among them as a directed labeled multigraph, where entities are represented as nodes and relations between them as labeled, directed edges. Accordingly, in the Semantic Web context a statement in a KG can be represented as a triple $(h,r,t)$, where $h$ is the head entity, $r$ the relation, and $t$ the tail entity, respectively. The connectivity among triples in KGs provides the basis for link prediction. 

Since the symbolic representations of KGs prohibit them from directly being incorporated in many machine learning tasks, recently many studies have proposed to embed entities and relations of a KG into low-dimensional vector spaces \cite{bordes2013translating,yang2014embedding,lin2015modeling,trouillon2016complex,schlichtkrull2018modeling,yan2019time}, which can be further unitized in multiple downstream tasks, e.g., the aforementioned link prediction. Along this line, there are two main branches \cite{wang2017knowledge}: (1) translation-based methods, which predict the existence of a triple by measuring the distance between the head entity and the tail entity after a translation enforced by the corresponding relation, such as TransE \cite{bordes2013translating}, TransD \cite{ji2015knowledge}, and TransR \cite{lin2015modeling} and (2) Semantic Matching Energy based methods, which measure the existence of a triple as the compatibility of two entities and their relation in latent vector space, e.g., RESCAL \cite{nickel2011three}, DistMult \cite{yang2014embedding}, ComplEx \cite{trouillon2016complex}. More recently, there are some other ideas. For example, rather than defining a relation as a translation from the  subject to the object, Sun et al. \cite{sun2018rotate} thought of a relation as a rotation from the subject to the object in the complex vector space and proposed RotatE, which was the first model that can handle symmetry/antisymmetry, inversion, and composition relations simulatenoiusly. Their experiments demonstrated the effectiveness of this assumption. More details about these methods can be found in Section \ref{sec:relatedwork}. 

Although there are multiple successful stories in both branches, these aforementioned models are all trained on individual triples independently regardless of their local neighborhood structures. Noticing this downside, Schlichtkrull et al. \cite{schlichtkrull2018modeling} state that explicitly modeling local structure can be an important supplement to help recover missing statements in KGs. Inspired by the success of graph convolutional networks (GCN) \cite{kipf2016semi} in modeling structured neighborhood information of unlabeled and undirected graphs with convolution operations, the authors proposed a GCN-based method to model knowledge graphs (R-GCN). 
In R-GCN, which is an encoder, the embedding of each entity is learned based on its up to $n$ degree neighboring entities by using $n$ graph convolution layers. Then the encoder is trained jointly with a task-specific decoder, e.g., a DistMult-like decoder, to predict links.

The experimental results of applying R-GCN demonstrate the importance of integrating neighborhood information in knowledge graph embedding models. 
R-GCN aims at learning entity embeddings even though it utilizes relation-specific weight matrices. The relation embeddings are learned in the task-specific decoder while the learned relation-specific matrices in the encoder are discarded.
Consequently, without a task-specific decoder for learning relation embeddings, R-GCN cannot directly support tasks such as link prediction.  Even if an extra decoder is available, the encoder-decoder framework runs into another problem of repeated introduction of relation-specific parameters in both the encoder side (relation-specific weight matrices) and the decoder side (relation embeddings). As a result, the number of parameters increases. 

To address the issue, we propose a novel model inspired by R-GCN \cite{schlichtkrull2018modeling}, a GCN-based knowledge graph encoder framework which can learn entity embeddings and relation embeddings simultaneously by performing relation-specific transformations from head entity embeddings to tail entity embeddings, hence called TransGCN. In principle, any presumed transformation assumption from the  subject to the object, such as translation assumption, rotation assumption, etc., can be exploited in the proposed framework. Take the translation assumption as an example, specifically in which translation operators acted by relations are  resorted to connect entities in a KG. The basic idea of TransGCN is illustrated in Figure 1. 
In such a scenario, TransGCN first translates the embeddings of 1-degree neighbors of one center entity $v_{i}$ with their specific relation embeddings. The resultant embeddings serve as the initial \textit{embedding estimations} of the center entity $v_{i}$. Then a convolutional operation is performed over these initially \textit{estimated} embeddings to derive a new embedding $\mathbf{v}_{i}^{\prime}$ for each $v_{i}$, which encodes local structural information of the center entity.
Similar to R-GCN, aggregated structural information and self-loop information of a node are combined for entity embedding updates. Moreover, we also define a novel relation embedding convolution process so that the entity and relation embeddings can be handled in a layer-based manner as GCNs do.

\begin{figure}[!h]
	\centering
	\includegraphics[width=0.2\paperwidth]{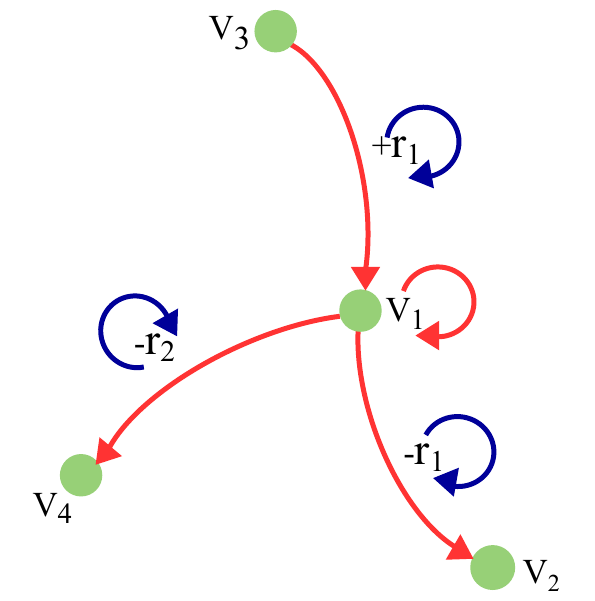}
	\label{fig:bi}
	\caption{The basic idea of TransGCN. Red lines show updating rules of an entity $v_1$, where the neighborhood information is aggregated from neighbors with corresponding relations (+/- denotes incoming/outgoing relations) and then is combined with self-loop information (red self-loop arrow). Blue lines disclose the update phases of relations, which are achieved simply by transforming the previous relation embeddings.}
\end{figure}

\textbf{The research contributions of our work are as follows:}
\begin{enumerate}
	\item We propose that transformation assumptions in which relations are assumed as transformation operators transforming the subject entity to the object entity can be utilized to convert a heterogeneous neighborhood in a KG into a homogeneous neighborhood, which can be readily utilized within a GCN-based framework.
	\item We develop a novel GCN-based knowledge graph encoder framework called TransGCN which can encode entity and relation embeddings simultaneously. Compared with R-GCN, this method has less parameters and can be directly used for link prediction.
	\item Based on the transformation assumptions behind TransE and RotatE, respectively, we instantiate our GCN framework. Experimental results on ~\fb~ and ~\wn~ show that two TransGCN models achieve substantial improvements against the state-of-the-art methods on both datasets.
\end{enumerate}
The paper is structured as follows. In Section \ref{sec:method} we elaborate on the main idea of the TransGCN framework. Experimental details on \fb\ and \wn\ are presented in Section \ref{sec:exp}. In Section \ref{sec:relatedwork}, we introduce two branches of learning methods on graphs. One is the classic translation-based models and the other are GCN-based approaches. Section \ref{sec:con} concludes this work and suggests future research directions.

\section{Proposed Architecture}  \label{sec:method}
R-GCN model does not learn relation embeddings and thereby would not be directly utilizable for link prediction without a decoder. Moreover, R-GCN model repeatedly introduces relation-specific parameters in both the encoder side and the decoder side, which results in an increase in the number of parameters. We argue that the  encoder alone for knowledge graph applications should encode entity and relation embeddings at the same time to reduce the number of parameters (thus helping alleviate the problem of overfitting) and thereby to improve training efficiency.

To address the issues, we propose a unified encoder framework based on GCN to learn entity and relation embeddings simultaneously, in which a presumed transformation assumption performed by relations is used to 
convert a heterogeneous neighborhood in a KG to a homogeneous one. This is subsequently used in a traditional GCN framework. Both entity embeddings and relation embeddings are learned in a convolutional layer-based manner.
A knowledge graph $\mathcal{G}=(\mathcal{V},\mathcal{E})$, where $\mathcal{V}$ is the set of nodes/entities and $\mathcal{E}$ is the set of labeled edges, contains statements in the form of a set of triples $(v_i,r_k,v_j) \in \mathcal{T}$, where $v_i$, $r_k$, and $v_j$ represent the head entity, the relation, and the tail entity, respectively. In the following, we use the bold text to refer to embeddings and we will use $(h,r,t)$ and $(v_i,r_k,v_j)$ interchangeably. 

\subsection{Handling a Heterogeneous Neighborhood in a KG}
\begin{figure}[!h]
	\centering	
	\includegraphics[width=0.3\textheight]{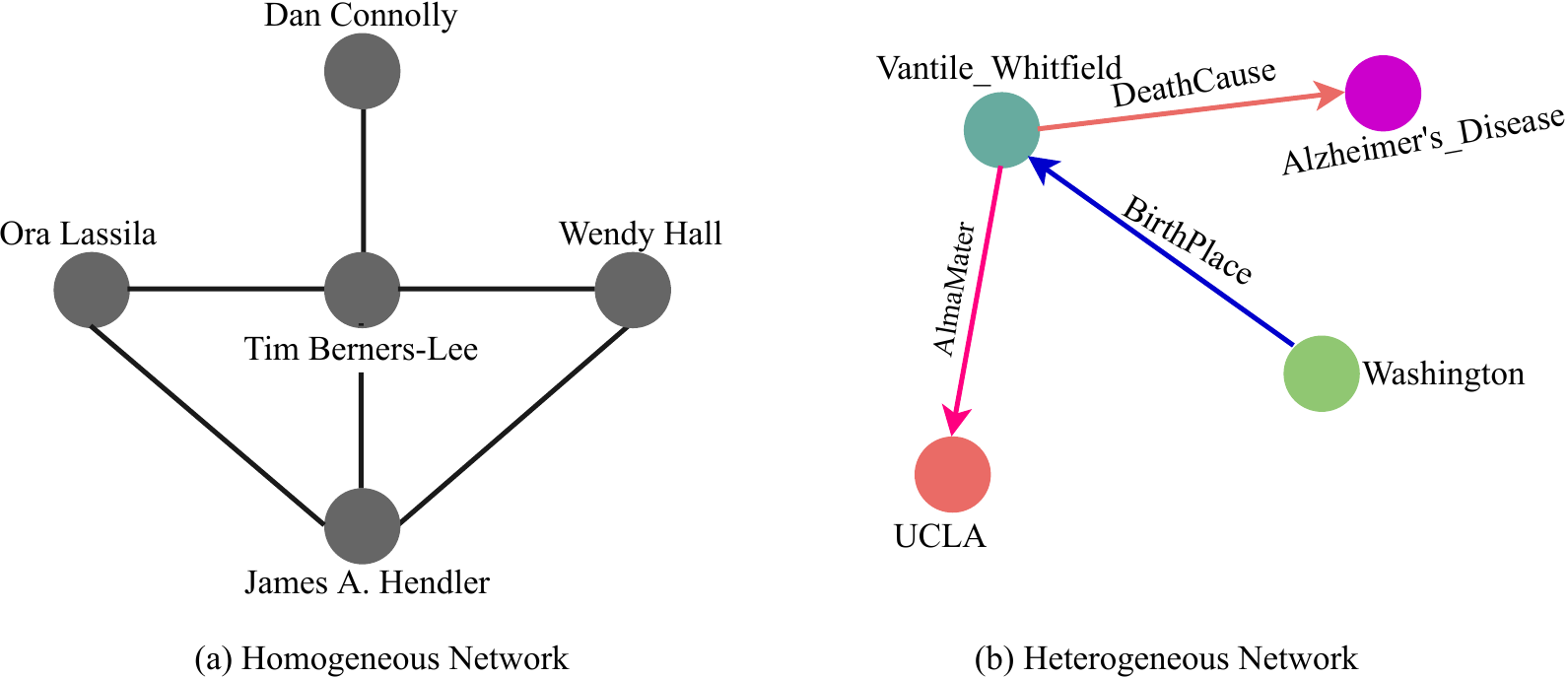}
	\caption{Neighbors in homogeneous and heterogeneous networks}
	\label{fig:neighbors}
		\vspace{-4mm}
\end{figure}

Traditional GCNs \cite{kipf2016semi} operate on an unlabeled undirected graph which consists of nodes of the same type and relations of the same type.
This means that each edge has the same semantics and the neighborhood of a node is homogeneous. We call this a \textit{homogeneous neighborhood}. Homogeneity makes it easier to aggregate the local neighborhood information around a node. For example, in an undirected unlabeled academic collaboration network shown in Figure \ref{fig:neighbors}(a), simply summing up information from \textit{Wendy\ Hall}, \textit{Dan\ Connolly}, \textit{Ora\ Lassila} and \textit{James A. Hendler} as messages transmitted to \textit{Tim\ Berners\-Lee} is reasonable. There is no need to consider the differences in messages since their relations in such a graph are the same.

However, in a knowledge graph such as shown in 
\ref{fig:neighbors}(b), using such oversimplified summations would be problematic. Neighboring entities are linked to the center entity via different relations in different directions. For instance, the relation \textit{DeathCause} is very different from the relation \textit{BirthPlace} and their directions to \textit{Vantile\_Whitfield} are pointing in opposite directions. We call this a \textit{heterogeneous neighborhood}. We argue that in order to make a KG be easily handled by a GCN-based framework, it is necessary to convert a heterogeneous neighborhood in a KG to a homogeneous one. In this work, we propose to approach this challenge by assuming relations in KGs are transformation operations which transform the head entity to the tail entity.

Common ways of transformation between two entities include translation, rotation, reflection, and so on. In any such a transformation, a statement in KGs can be interpreted as that the head entity is transformed to the tail entity by a relation. More specifically, the tail entity in a statement may be the head entity after being translated/rotated. Accordingly, the embedding of a tail entity can be estimated by the head entity after a relation-specific transformation operation. For ease of  generality, a statement $<h,r,t>$ following a transformation assumption can be written as:

\begin{equation}
\begin{cases}
\mathbf{t}=\mathbf{h}\circ\mathbf{r}\\
\mathbf{h}=\mathbf{t}\star\mathbf{r}
\end{cases}
\label{eq:transformation}
\end{equation}
where $\star$ and $\circ$ are defined as two transformation operators, which vary from different assumptions. We will specify them later in \ref{transE-gcn} and \ref{rotetE-gcn}. The diversity of relation types and the direction of relations are two main characteristics of heterogeneity of heterogeneous graphs, e.g. KGs. Obviously, in this equation, the fact that each relation type is encoded differently takes care of the diversity of relation types and the transformation operators are usually specially designed to address the relation direction.

Based on the transformation assumption, we define the estimations of a central entity derived from connected entities with corresponding relations as the embeddings of neighbors of the entity. Take TransE as an example. Given an entity $v_i$ with an outgoing triple $(v_i,r_k,v_j)$, we define the estimation ($\bf{v}_j-\bf{r}_k$) of $v_i$ based on $(v_i,r_k,v_j)$ as the embedding of one neighbor of $v_i$. Similarly, for an incoming triple $(v_l,r_m,v_i)$ of $v_i$, the embedding of another neighbor is $\bf{v}_l+\bf{r}_m$, which is another estimation of the central entity $v_i$.  
More concretely, in Figure \ref{fig:neighbors}(b), the estimations of the entity $v_{Vantile\_Whitfield}$ from incoming triples can be expressed as $\{\mathbf{v}_{Washington}+\mathbf{r}_{BirthPlace}\}$, while embedding estimations from outgoing triples can be expressed as $\{ \mathbf{v}_{UCLA}-\mathbf{r}_{AlmaMater},\\
\mathbf{v}_{Alzheimer's\_Disease}-\mathbf{r}_{DeathCause}\}$.

Formally, under any transformation assumption, the embedding estimations of an entity $v_i$ can be shown as follows:
\begin{align}\label{eq:neighbor}
\centering
\begin{split}
\mathcal{T}(v_i)&=\mathcal{T}_{in}(v_i)\cup \mathcal{T}_{out}(v_i) 
\\
\mathcal{T}_{in}(v_i)&=\{(v_j,r_k,v_i)\ |\ \forall v_j,r_k\ (v_j,r_k,v_i)\in \mathcal{T}\}
\\
\mathcal{N}_{in}(v_i)&=\{\mathbf{v_j}\circ\mathbf{r_k}\ |\ \forall (v_j,r_k,v_i)\in \mathcal{T}_{in}(v_i)\}
\\
\mathcal{T}_{out}(v_i)&=\{(v_i,r_k,v_j)\ |\ \forall v_j,r_k\ (v_i,r_k,v_j)\in \mathcal{T}\}
\\
\mathcal{N}_{out}(v_i)&=\{\mathbf{v_j}\star\mathbf{r_k}\ |\ \forall (v_i,r_k,v_j)\in \mathcal{T}_{out}(v_i)\}
\\
\end{split}
\end{align}
where $\mathcal{T}(v_i)$ denotes all the triples associated with $v_i$, consisting of $\mathcal{T}_{in}(v_i)$ as incoming triples and $\mathcal{T}_{out}(v_i)$ as outgoing triples, and $\mathcal{N}_{in}(v_i)$ and $\mathcal{N}_{out}(v_i)$ both are the sets of the estimated embeddings derived from incoming and outgoing neighbors, respectively. 

After these transformation operations along different triple paths, the resultant estimated embeddings for the center entity should have the same semantics to the true center entity, by which the heterogeneous neighborhood in a KG is converted to a homogeneous one that can be easily handled by the GCN framework.

\subsection{Model Formulation}
Our model can be regarded as an extension of R-GCN \cite{schlichtkrull2018modeling}. In the following, we introduce how our model learns entity and relation embeddings at the same time. Like other existing GCN models \cite{kearnes2016molecular,schlichtkrull2018modeling}, our model can be formulated as a special case of Message Passing Neural Networks (MPNN)~\cite{gilmer2017neural}, which provide a general framework for supervised/semi-supervised learning on graphs. 

In general, MPNN defines two phases: a message passing phase for nodes and a readout phase for the whole graph. Since in this paper we care about nodes and relations instead of the whole graph, we focus only on the message passing phase. 
Basically, this message passing phase of a node is executed $L$ times to aggregate multi-hop neighborhood information and is composed of message passing functions $M^{(l)}$ and node update functions $U^{(l)}$, where $l$ denotes the l-th hidden layer. $M^{(l)}$ mainly aggregates messages from local neighbors, while $U^{(l)}$ combines $M^{(l)}$ with self-loop information in the previous step. 
Both of these two functions are differentiable. In addition, Gilmer et al.~\cite{gilmer2017neural} indicated that one could also learn edge features by introducing similar functions for all the edges in a graph, but so far only Kearnes et al.~\cite{kearnes2016molecular} have implemented this idea. To fit it into our task, we instantiate $M^{(l)}$ and $U^{(l)}$ for message propagation and entity embedding update for each entity $v_i$, and additionally introduce the update rule for a relation. 
{\small 
	\begin{align}
	\centering
	\label{eq:messagepassing}
	\begin{split}
	\mathbf{m}_i^{(l+1)}&=\sum\limits_{(v_j,r_k,v_i) \in \mathcal{T}(v_i)}M^{(l)}(\mathbf{v}_i^{(l)},\mathbf{v}_j^{(l)},\mathbf{r}_k^{(l)}) \\
	&=\frac{1}{c_i}\mathbf{W}_0^{(l)}\big(\sum\limits_{(v_j,r_k,v_i) \in \mathcal{T}_{in}(v_i)} (\mathbf{v}_j^{(l)} \circ \mathbf{r}_k^{(l)})\\
	&+\sum\limits_{(v_i,r_k,v_j) \in \mathcal{T}_{out}(v_i)} (\mathbf{v}_j^{(l)} \star \mathbf{r}_k^{(l)})\big)
	\end{split}
	\end{align}
}

\begin{equation}
\label{eq:entityupdate}
\mathbf{v}_i^{(l+1)}=U^{(l)}(\mathbf{m}_i^{(l+1)},\mathbf{v}_i^{(l)})=\sigma(\mathbf{m}_i^{(l+1)}+\mathbf{v}_i^{(l)})
\end{equation}
where $\mathbf{v}_i^{(l)} \in \mathbb{C}^{d^{(l)}}$ denotes the hidden representation of entity $v_i$ in the $l$-th layer with a dimensionality of $d^{(l)}$. $\mathbf{W}_0^{(l)} \in \mathbb{C}^{d^{(l+1)}\times d^{(l)}}$ is a layer-specific matrix. $\mathcal{T}_{in}$ and $\mathcal{T}_{out}$ are defined in Eq.~\ref{eq:neighbor}. $c_i$ is an entity-related normalization constant that could be the total degree of $v_i$. $\sigma$ is the activation function, e.g., $ReLU$.

\begin{figure}[!h]
	\centering
	\setlength{\unitlength}{0.1\textwidth}	
	\includegraphics[width=0.3\textheight]{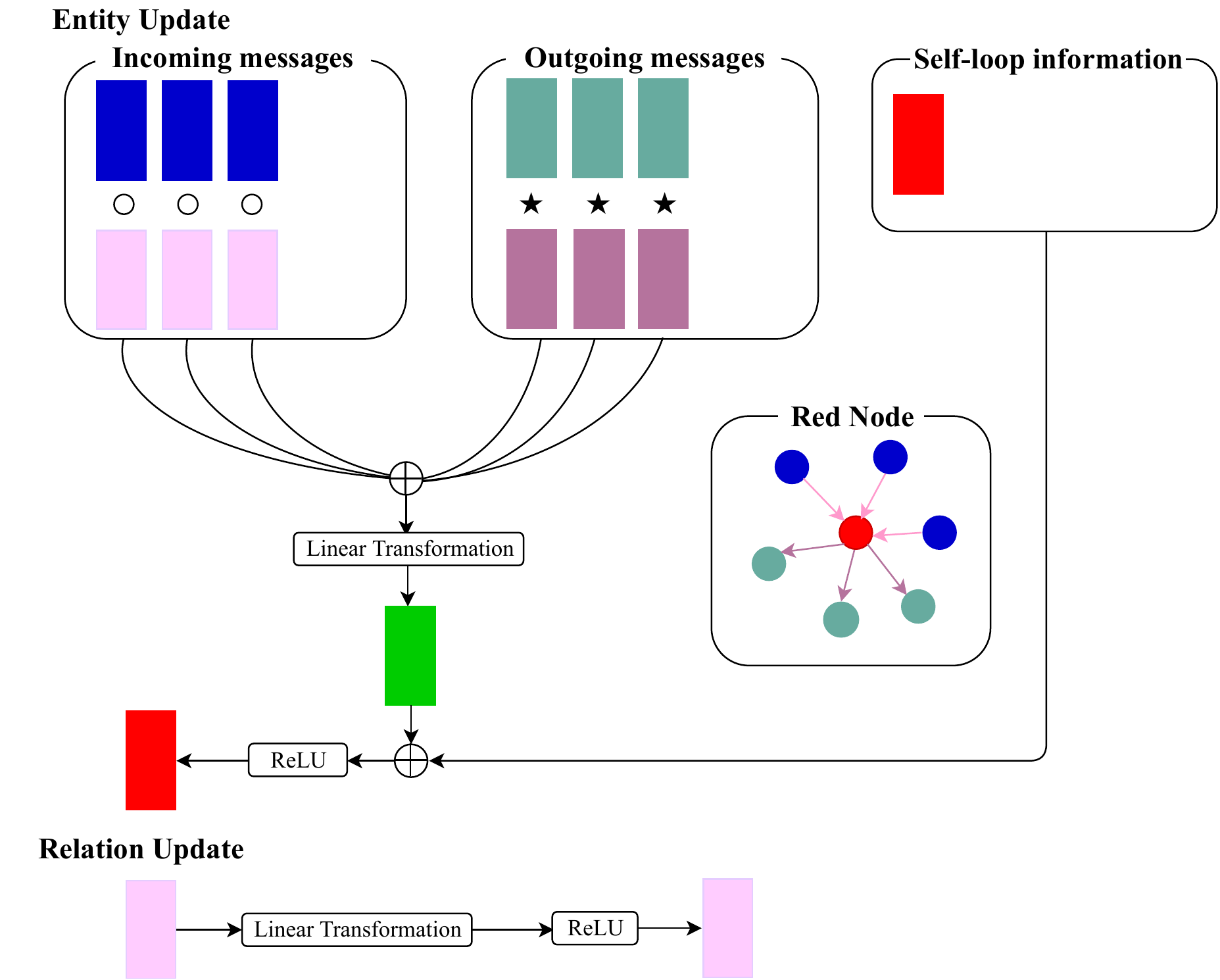}
	\caption{Diagram for the update of entity/node and relation/edge in the proposed TransGCN model.}
	\label{fig:diagram}
\end{figure}

Basically, there are two terms in Eq.~\ref{eq:messagepassing} that are used to encode local structural information for entity update representing messages from incoming relations and outgoing relations, respectively. The messages from incoming/outgoing relations are first accumulated by an element-wise summation and then are passed through a linear transformation. Then in the next step (Eq.~\ref{eq:entityupdate}), these messages are combined with self-loop information by simply adding them up to update entities. This idea is inspired by the skip-connections in ResNet~\cite{He_2016_CVPR} so that our model can perform at least as well as the simple transformation-based model instantiated in this framework. Figure \ref{fig:diagram} illustrates the computation graph for an entity. Typically, Eq.~\ref{eq:messagepassing} considers the first-order neighbors of entities. One could simply stack multiple layers to allow for multi-hop neighbors.
In addition, we realize that every update of entity embeddings in Eq.~\ref{eq:messagepassing} and Eq.~\ref{eq:entityupdate} may transform the original vector space. Consequently, the relationships between relations and entities would be affected, which makes it impossible to perform presumed transformation operations between them in the next layer. To address this problem, instead of applying a similar message passing mechanism for relations as in Eq.~\ref{eq:messagepassing} and Eq.~\ref{eq:entityupdate}, for ease of efficiency, we introduce a transformation matrix $\mathbf{W}_1^{(l)}$ operated on relation embeddings for each layer. We assume that the introduced matrix can project relation embeddings into a vector space that has the same relation to the new entity vector space as they have before. Note that this is a soft restriction on the vector space; one could choose other more strict restrictions as well. For example, enforce constraints on the basis vectors of entities and relations so that these two vector spaces are ensured to be the same. Following the soft restriction, the update rule of relations in each layer is formed as follows:
\begin{equation}
\label{eq:relationupdate}
\mathbf{r}_k^{(l+1)} = \sigma(\mathbb{W}_1^{(l)}\mathbf{r}_k^{(l)})
\end{equation}
where $\mathbf{r}_k^{(l)} \in \mathbb{C}^{d^{(l)}}$ is the hidden state of relation $r_k$ in the $l$-th layer with a dimensionality of $d^{(l)}$. $\mathbf{W}_1^{(l)} \in \mathbb{C}^{d^{(l+1)}\times d^{(l)}}$ is a linear transformation across relations in the $l$-th layer. 

In the following subsections, we instantiate our TransGCN framework by using two different transformation assumptions. One is based on the translation assumption and TransE is selected owe to its simplicity and popularity, while the other follows the rotation assumption and RotatE is chosen to achieve this assumption.

\subsection{TransE-GCN model} \label{transE-gcn}
Under the translation assumption, the relation is assumed to serve as a translation from the head entity to the tail entity. For an entity $v_i$, Eq.~\ref{eq:transformation} can be instantiated as follows:
\begin{equation}
\mathbf{v}_i^{(l)}=\begin{cases}
\mathbf{v}_j^{(l)}+\mathbf{r}_k^{(l)}, ~(v_j, r_k, v_i) \in \mathcal{T}_{in}(v_i)\\
\mathbf{v}_j^{(l)}-\mathbf{r}_k^{(l)}, ~(v_i,r_k,v_j) \in \mathcal{T}_{out}(v_i)
\end{cases}
\label{eq:TransE}
\end{equation}
where $\circ$ and $\star$ are $+$ and $-$, respectively.

Like in TransE, the score function for a triple $(v_i,r_k,v_j)$ is defined according to:
\begin{equation}
f_{r_{k}}(v_i,v_j)=-\norm{\mathbf{v}_i^{(l)}+\mathbf{r}_k^{(l)}-\mathbf{v}_j^{(l)}}
\label{eq:trans_score}
\end{equation}
where $\mathbf{v}_i^{(l)}$, $\mathbf{r}_k^{(l)}$ and $\mathbf{v}_j^{(l)}$ are the embeddings of $v_i$, $r_k$ and $v_j$ in the last layer, respectively.

Similar to previous studies \cite{wang2014knowledge}, this model is trained with negative sampling. For each existing triple in a KG, a certain number of negative samples (e.g., one positive triple with 10 negative samples) are constructed by replacing either the head entity or the tail entity randomly. Positive samples are expected to have high scores while negative samples are expected to have low scores. A margin-based ranking function is written as the loss function for training:
\small
\begin{equation}
\mathcal{L}=\sum\limits_{(v_i,r_k,v_j)\in\mathcal{T}} \sum\limits_{(v_i^{\prime},r_k,v_j^{\prime})\in\mathcal{T^{\prime}}} \max(0,-f_{r_{k}}(v_i,v_j)+f_{r_{k}}(v_i^{\prime},v_j^{\prime})+\gamma)
\end{equation}
\normalsize
where $\max(a,b)$ is used to obtain the maximum between $a$ and $b$, $\gamma$ is the margin, $\mathcal{T}$ is the set of observed triples in a KG, and $\mathcal{T}^{\prime}$ is the set of negative samples associated with the positive sample $(v_i,r_k,v_j)$. It is noteworthy that in this implementation, all the embeddings are  in the real vector space.

\subsection{RotatE-GCN model} \label{rotetE-gcn}
Another assumption recently explored in knowledge graph embedding  is rotation.  Sun et al. assumed that the tail entity is derived from the head entity after being rotated performed by a relation in the complex vector space \cite{sun2018rotate}. Accordingly, we can formalize the neighbors of an entity $v_i$:

\begin{equation}
\mathbf{v}_i^{(l)}=\begin{cases}
\mathbf{v}_j^{(l)}\odot\mathbf{r}_k^{(l)}, ~(v_j, r_k, v_i) \in \mathcal{T}_{in}(v_i)\\
\mathbf{v}_j^{(l)}\oplus\mathbf{r}_k^{(l)}=\mathbf{v}_j^{(l)}\odot \overline{r}_k^{(l)}, ~(v_i,r_k,v_j) \in \mathcal{T}_{out}(v_i)
\end{cases}
\label{eq:RotatE}
\end{equation}

where $\circ$ and $\star$ are $\odot$ and $\oplus$, respectively. More specifically, $\odot$ is the element-wise product in the complex space and $\overline{r}_k$ is the complex conjugate  of $r_k$. $\mid r_i \mid=1$. Note that here the existence of $r_k$ and $\overline{r}_k$ rather than different transformation operators guarantees the relation direction is considered naturally.

Similarly, the distance function serves as the score function:
\begin{equation}
f_{r_{k}}(v_i,v_j)=-\norm{\mathbf{v}_i^{(l)}\odot\mathbf{r}_k^{(l)}-\mathbf{v}_j^{(l)}}
\label{eq:rotate_score}
\end{equation}

To keep consistent with RotatE, we adopt self-adversarial negative sampling to train the model rather than vanilla negative sampling. The main argument of self-adversarial negative sampling is that negative triples should have different probabilities of being drawn as training continues, e.g. many triples may be obviously false, thus not contributing any meaningful information. Therefore, a probability distribution $p$ is used to draw negative samples according to the current embedding model.
\small 
\begin{equation}
	p(v_i^{\prime}, r_k, v_j^{\prime}|{(v_i,r_k,v_j)})=\frac{exp\big(\alpha f_{r_k}(v_i^{\prime}, v_j^{\prime})\big)}{\sum_{(v_i^{\prime\prime}, r_k, v_j^{\prime\prime}) \in {\mathcal{T^{\prime}}}}exp\big(\alpha f_{r_k}(v_i^{\prime\prime}, v_j^{\prime\prime})\big)}
\end{equation}   
\normalsize
where $\alpha$ is a constant which controls the temperature of sampling and $\sigma$ is the sigmoid function.

Then the above probability of a negative sample is treated as the weight of the sample to help construct the loss function. For a positive sample $(v_i, r_k, v_j)$, the loss function can be written as follows:
\begin{multline}
\mathcal{L}=-log\big(\sigma(\gamma+f_{r_{k}}(v_i,v_j))\big)\\
-\sum\limits_{(v_i^{\prime},r_k,v_j^{\prime})\in \mathcal{T}^{\prime}}p(v_i^{\prime}, r_k, v_j^{\prime})log\big(\sigma(-f_{r_{k}}(v_i^{\prime},v_j^{\prime})-\gamma)\big)
\label{eq:rotate_loss}
\end{multline}
where all the embeddings are in the complex vector space.

\section{Experiment} \label{sec:exp}
To test the performance of our models, we evaluate our TransGCN models on the task of link prediction on two datasets: \fb\ and \wn.
\vspace{-2mm}
\subsection{Datasets}
In previous studies, the performance of link prediction methods was commonly evaluated on two datasets, namely FB15K from Freebase and WN18 from WordNet. However, there are inverse triples in both training and testing data, resulting in methods showing better performance on these datasets by means of memorizing these affected triples rather than having a better ability of prediction. Therefore, we use the two filtered data sets: \fb~ and \wn, proposed in ~\cite{toutanova2015observed} and  ~\cite{dettmers2018conve}, respectively, in which all the inverse triplet pairs were removed. These two datasets have been shown to be more challenging for models to perform link prediction~\cite{schlichtkrull2018modeling}. Table \ref{tb:basic} shows basic statistics for these two datasets. 

\begin{table}
	\centering
	\caption{Basic statistics of \fb\ and \wn.}
	\label{tb:basic}
	\vspace{-4mm}
	\scalebox{0.8}{
	\begin{tabular}{>{\centering\arraybackslash}m{1.25in}  *2{>{\centering\arraybackslash}m{.85in}}} \toprule[1.25pt]
		\bf Dataset& \bf FB15k-237& \bf WN18RR \\
		\hline
		\bf Entities &14,541 & 40,943\\
		\bf Relations &237 &11\\
		\bf Training triples&272,115&86,835\\
		\bf Validation triples&17,535&3,034\\
		\bf Test triples&20,466&3,134\\
		\bottomrule[0.85pt]
	\end{tabular}}
		\vspace{-4mm}
\end{table}
\vspace{-2mm}
\subsection{Experiment Setup}
\paragraph{Evaluation metrics.} In the testing phase, for each triple, we replace the head entity with all other entities in current KG, and calculate scores for those replaced triples and the original triple using the scoring function specified in section \ref{sec:method}. Since some of the replaced triples might also appear in either training, validation or test set, we then filter these triples out and produce a filtered ranking which we denote as the \textit{filtered} setting.  Then those triples are ranked in a descending order of scores and the rank of the correct triple in this ranking list is used for evaluation. The whole procedure is repeated while replacing the tail entity instead of the head entity. Following previous studies \cite{bordes2013translating}, we adopt Mean Reciprocal Rank (MRR) and Hits@k as evaluation metrics. 
We report filtered MRR scores as well as Hits at 1, 3, and 10 for the \textit{filtered} setting. For all the metrics, higher values mean better performance.

\paragraph{Baselines.} Six baselines  ($TransE$, $DistMult$, $ComplEx$, $R-GCN$, $ConvE$ and $RotatE$) are selected for the evaluation. $TransE$ is a standard translation-based model, which is simple but performs well on most datasets. This model is wrapped in our TransE-GCN model to achieve the conversion from heterogeneous neighbors to homogeneous neighbors. $DistMult$, as a factorization model, also shows promising performance on standard datasets. Furthermore, our model is compared with $ComplEx$~\cite{trouillon2016complex}, one powerful state-of-the-art model for link prediction, and $R-GCN$ \cite{schlichtkrull2018modeling}, a strong baseline of modeling directed labeled graph. $ConvE$ uses a multi-layer convolutional network to model the iterations between entities and relations\cite{dettmers2018conve}. $RotatE$ is the most recent KGE model, which is built on the rotation assumption ~\cite{sun2018rotate}. This model is exploited in our RotatE-GCN model to derive homogeneous neighbors. 

\paragraph{Implementation details.} To optimize our TransGCN models, we used the Adam optimizer~\cite{kingma2014adam} and fixed the learning rate $\lambda =0.001$. The best parameters were selected when filtered MRR achieved the best performance on respective validation sets. First, for both models, the embeddings of entities and relations produced by these two base models, i.e. TransE and RotatE, were used to initialize the embeddings needed in our models. For TransE, the embeddings pretrained by Nguyen et. al in \cite{Nguyen2018} were utilized. Then, the number of layers in GCN was selected by comparing the experimental results on validation set. Finally, $L=1$ was the best choice for both datasets. For RotatE, we trained this model by using the implementation provided by the authors to gain initial embeddings of entities and relations.  Then most of the hyperparameter values of RotatE remained unchanged except that we ignored the batch size, since in GCN the batch size is achieved by setting graph batch size, which we leave as default. The only tuned parameter was the number of layers $L \in \{1,2\}$. Finally, the best parameter settings in our experiment are $ L=2$ on \fb\ and $L=1$ on \wn.

%


\vspace{-4mm}
\subsection{Results} \label{sec:results}
\paragraph{Main Results.} The results for both datasets are reported in Table \ref{tb:results}. Results on the  baseline models $DistMult$, $TransE$, $ComplEx$, $ConvE$, and $RotatE$ are taken from \cite{sun2018rotate}, and R-GCN's results are taken from \cite{schlichtkrull2018modeling}.

\begin{table*}[]
	\caption{Prediction results of different models on \fb~ and \wn~}
	\label{tb:results}
	\scalebox{0.8}{
	\begin{tabular}{|c|c|ccc|c|ccc|}
		\hline
		\multirow{3}{*}{} & \multicolumn{4}{c|}{FB15K-237}                                                     & \multicolumn{4}{c|}{WN18RR}                                                          \\ \cline{2-9} 
	 & MRR(Filtered)        & Hit@1              & Hit@3              &Hit@10                  & MRR(Filtered)        &Hit@1              &Hit@3              &Hit@10             \\ \hline
		DistMult              & 0.241        & 0.155          & 0.263          & 0.43               & 0.39          & 0.44         & 0.49          & 0.447          \\
		TransE            & 0.294          & -          		& -         		& 0.465             & 0.226           & -      & -        &0.501 \\ 
		TransE-GCN   & 0.315          &0.229    		&0.324      	&0.477     			&  0.233		 &0.203		&0.338			&0.508	\\ \hline
		ComplEx   & 0.247          & 0.158          & 0.275          & 0.428          			& 0.44 		&0.41 			& 0.46          & 0.51         \\
		R-GCN       & 0.248          & 0.153          & 0.258          & 0.417          		& -           & -          & -          & -          \\ 
		ConvE		&0.325 			&0.237				&0.356 			&0.501 					&0.43		&0.40		& 0.44       &0.52  \\
		RotatE        & 0.338           & 0.241           & 0.375        & 0.533        		&0.476           & 0.428          & 0.492          & 0.571   \\ 
		RotatE-GCN   & \textbf{0.356} & \textbf{0.252} & \textbf{0.388} & \textbf{0.555}  &\textbf{0.485}      & \textbf{0.438}       & \textbf{0.51} & \textbf{0.578} \\ \hline
	\end{tabular}}
\end{table*}

In Table \ref{tb:results}, one important observation is that our TransE-GCN model and RotatE-GCN model both outperformed their base models, i.e. TransE and RotatE, on both datasets in terms of all the metrics by noticeable margins, which demonstrates the effectiveness of our proposed framework. Besides, the improvements restate the significance of explicitly incorporating local structural information in knowledge graph embedding learning. Moreover, compared with all the other baselines, the RotatE-GCN model was consistently better while the TransE-GCN model performed differently on the two datasets. To be specific, TransE-GCN performed better than ComplEx on ~\fb~ while worse on the other dataset. This can be interpreted by the difference between TransE and ComplEx that TransE is not good at dealing with relation types except 1-to-1 relations, as pointed out by researchers before. During the training process, for each triple (h,r,t), TransE enforces $\mathbf{h}+\mathbf{r}$ to be as close as possible to $\mathbf{t}$, which would be problematic when dealing with 1-to-N, N-to-1, and N-to-N relations. For example, given a 1-to-N relation $r$, we have two triples $(h,r,t_1)$ and $(h,r,t_2)$. If $\bf{h}+\bf{r}=\bf{t}$ holds, $\mathbf{t}_1$ and $\mathbf{t}_2$ should have the same vector representations. To meet this requirement, $\bf{h}+\bf{r}$ is close to the center of all the positive tails $\bf{t}$ at the end of training instead of a particular tail (which may be the correct prediction). Therefore, the performance of TransE dropped extremely on ~\wn, where there are four times more entities but 20 times less relations than those in ~\fb. The superior performance of  RotatE-GCN model over TransE-GCN model indirectly showed the importance of a base model used in our framework.
\vspace{-4mm}
\paragraph{Comparison with R-GCN} It is necessary to elaborate on the comparison between our models and the R-GCN model that inspired our work. The experimental results showed that our models (TransE-GCN, RotatE-GCN model) both consistently yielded better results with improvements of 10.8\% and 6.7\% in terms of $MRR(Filtered)$ on FB15K-237, respectively. We believe the improvements are attributed to two reasons. First, thanks to the idea of converting heterogeneous neighbors into homogeneous neighbors in KGs, proposed in this paper, it successfully captured both local structural information by considering entities and relations in the neighborhood and semantic information residing within transformation operators. Besides, by doing so, relations in a KG were just modeled once and simultaneously with entities, and relation-specific matrices in R-GCN being replaced by shared matrices potentially facilitated the encoding of more complex latent information. Thus, fewer parameters were needed to learn in our models, which helps alleviate the problem of overfitting. In total, our TransE-GCN model has $((B-1) \times L\times d^{2}+2\times B \times R \times L)$ fewer parameters than R-GCN in terms of basis decomposition regularization and $(2 \times B \times R \times L \times(\frac{d}{B})^2-L\times d^{2})$ fewer parameters in terms of block-diagonal decomposition, where $B$ denotes the number of basis matrices, $L$ denotes the number of layers, $d$ denotes the dimension of a hidden layer, and $R$ denotes the number of relations.  As for our RotatE-GCN model, we followed the implementation proposed by \cite{sun2018rotate}. They used real numbers to express complex numbers by treating the first half dimensions of entity embeddings as the real part and the last half as the imaginary part. Therefore, the dimensions of entities are doubled in the complex vector space. Finally, our RotatE-GCN model has $((B-5) \times L\times d^{2}+2\times B \times R \times L-E \times d)$ fewer parameters than R-GCN (basis decomposition) and $(2 \times B \times R \times L \times(\frac{d}{B})^2-5\times L\times d^{2}-E\times d)$ fewer parameters than R-GCN (block-diagonal decomposition), respectively, in which $E$ is the number of entities.

\begin{figure}[!h]
	\centering	
	\includegraphics[width=0.2\textheight]{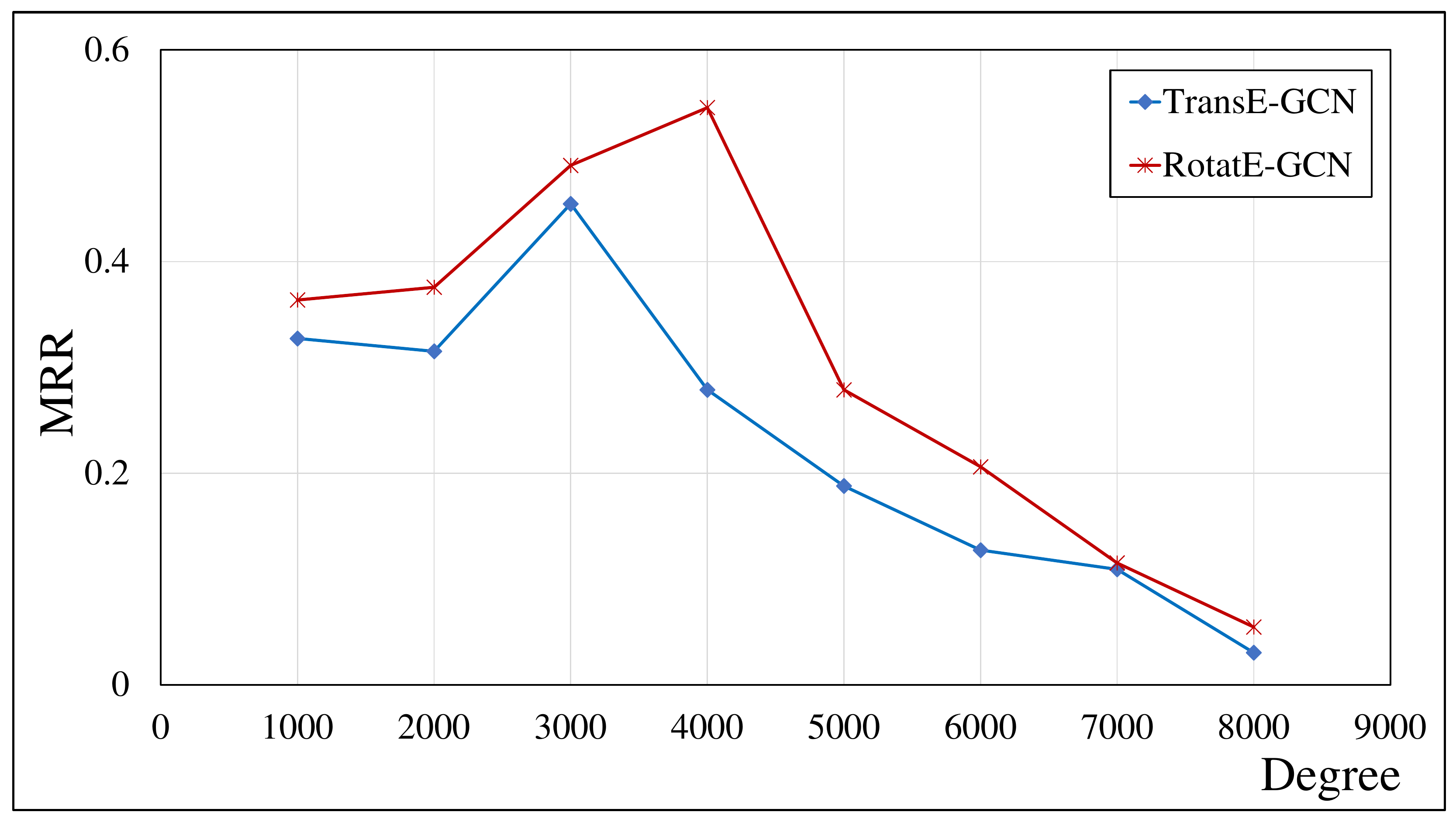}
	\caption{MRR for TransE-GCN and RotatE-GCN on FB15k-237 validation set with the entity degree}
	\label{fig:degree}
\end{figure}

\paragraph{Performance on Entities of different degrees}  Figure \ref{fig:degree} depicts the performance of our models on FB15K-237 validation set as functions of the entity degree. It can be observed that in the beginning the performance of both models increased a lot with the increasing size of neighborhood, while after a threshold, it dropped significantly. We believe it showed that a few neighbors were only able to provide limited local structural information, thus leading to poor performance; by contrast, too many neighbors brought too much mixed information, which made models hard to optimize. In the future, more work should focus on how to deal with these two extreme conditions.

\begin{table}[!h]
	\caption{Prediction results of our models on \fb~ in terms of different hops}
	\label{tb:neighbor_analysis}
	\vspace{-4mm}
		\begin{tabular}{c c c}
			\hline
			& MRR   & Hit@10 \\ \hline
			\bf{TransE-GCN-1} & 0.315 & 0.474  \\
			TransE-GCN-2 & 0.297 & 0.453  \\
			TransE-GCN-3 & 0.273 & 0.421   \\ \hline
			RotatE-GCN-1 & 0.347 & 0.546  \\
			\bf{RotatE-GCN-2} & 0.356 & 0.555  \\
			RotatE-GCN-3 & 0.331 & 0.525  \\ \hline
	\end{tabular}
	\vspace{-4mm}
\end{table}

\paragraph{Analysis of multi-hop neighbors} Table \ref{tb:neighbor_analysis} describes the prediction performance of our two models on ~\fb~ in terms of multi-hop neighbors, namely 1-hop, 2-hop and 3-hop neighbors. TransE-GCN model favored 1-hop neighbors while RotatE was able to leverage more neighborhood information. The difference lies in that RotatE has a stronger ability to deal with complex relations and capture more accurate entity and relation information. But both models performed the worst when 3-hop neighbors were considered, which were even worse than the base models. We think this may be caused by spectral convolutional filters, since it has been proven to have a smooth effect that could dilute the useful information\cite{li2018deeper}.

\section{Related Work} \label{sec:relatedwork}
Here we review previous work as it relates to our model.
\subsection{Transformation-based Models}
Until now, there exist two transformation assumptions in the literature - Translation and Rotation. A multitude of studies have explored these assumptions to achieve knowledge graph embedding learning for downstream tasks.

Translation-based models, also known as translational distance models, employed distance-based functions to model entities and relations in a KG. The key idea behind this kind of models is that for a positive triple $(h,r,t)$, the head entity should be as close as possible to the tail entity through the relation, serving as a translation. 

The most representative model is TransE \cite{bordes2013translating} because of its simplicity and efficiency. It encodes the observed triples in a KG and projects entities and relations into the same vector space. TransE directly implemented the vanilla idea of translation, which enforces $\mathbf{h}+\mathbf{r}=\mathbf{t}$ when $(h,r,t)$ holds. However, Wang et al.~\cite{wang2014knowledge} argued that TransE cannot deal with N-to-1, 1-to-N and N-to-N relations and proposed a new model called TransH, which introduces a hyperplane $H_r$ for each relation and requires that the projected head entity $h^{\prime}$ on $H_r$ should be close to the projected tail entity $t^{\prime}$ on $H_r$ after a translation $r$. TransR~\cite{lin2015modeling} follows a similar idea, but it introduced relation-specific translation spaces. In such a way, relations and entities can be represented in respective vector spaces. TransD~\cite{ji2015knowledge} and TranSparse~\cite{ji2016knowledge} are two other alternative approaches to simplifying TransR. In addition, another branch of improving TransE is to relax the strict restriction of $\bf{h}+\bf{r}=\bf{t}$, such as TransF~\cite{feng2016knowledge}. For example, TransM assigned each triple with a relation-specific weight $\theta_r$, and redefined the scoring function as $f_r(h,r)=-\theta_r\norm{\bf{h}+\bf{r}-\bf{t}}$. For a comprehensive review of these methods, please refer to \cite{wang2017knowledge}. Our TransE-GCN model was based on this translation assumption and TransE from the first branch was exploited for carrying out translation operations.

Rotation assumption was recently exploited by Sun et al.\cite{sun2018rotate}. Motivated by the Euler's identify that indicates a rotation in the complex plane can be achieved a unitary complex number,  Sun et al. proposed a RotatE model, which projected both entities and relations into the complex vector space and treated each relation as a rotation from the head entity and the tail entity. The most attractive characteristics of RotatE is its ability to model and to infer multiple relation patterns, including symmetry/antisymmetry, inversion and composition. This model also adopted a distance-based score function to evaluate the compatibility of two entities and their relations, as shown in ~\ref{eq:rotate_score}. Robust experimental results on benchmark datasets demonstrated the effectiveness of RotatE.

\subsection{Graph Convolutional Networks}
Our TransGCN framework is primarily motivated by plenty of works on modeling large-scale graph data using GCNs \cite{kipf2016semi}. Generally, GCN can be classified into: (1) spectral-based approaches, which introduce spatial filters from the graph signal processing perspective~\cite{shuman2013emerging, li2018adaptive}; (2) spatial-based approaches, which simply interpret a graph convolutional operation as aggregating information from neighbors~\cite{gilmer2017neural,hamilton2017inductive,gao2018large}. Although spectral-based methods seem appealing in that they can be supported by the spectral graph theory, in practice spatial-based methods perform better in terms of efficiency, generality and flexibility~\cite{wu2019comprehensive}. 

Interestingly, Kipf and Welling~\cite{kipf2016semi} discovered that when approximated by the $1^{st}$ order Chebyshev polynomials, the graph convolution is localized in space. That is, to some degree spatial-based approaches are the same as spectral-based approaches. Based on this, they introduced a simple but efficient message propagation rule conditioned on nodes and adjacency matrix of a graph for the semi-supervised node classification task.

To extend the GCN model~\cite{kipf2016semi} to directed labeled graphs\cite{schlichtkrull2018modeling} proposed an R-GCN model, which is the first work that applied the GCN framework to knowledge graphs for link prediction. The main contribution of this work lies in the introduction of relation-specific weight matrices in each layer of a neural network such that relation-specific messages can be propagated over graphs for entity update. The message propagation method for node $v_i$ is defined as follows:

\begin{equation}
{\scriptsize }
\mathbf{v}_i^{(l+1)}=\sigma\big(\sum\limits_{r \in \mathcal{R}}\sum\limits_{j \in \mathcal{N}_i^{r}}\frac{1}{c_{i,r}}W_r^{(l)}\mathbf{v}_j^{(l)}+W_0^{(l)}\mathbf{v}_i^{(l)}\big)
\end{equation}
where $W_r^{(l)}$ denotes a relation-specific weight matrix in the $l$-th layer, $W_0^{(l)}$ another layer-specific weight matrix, $\mathcal{R}$ the set of relation types and $\mathcal{N}_i^{r}$ the set of neighbors of node $v_i$ in terms of relation $r$.

To perform the task of link prediction, R-GCN, as an encoder, must cooperate with a decoder, such as DistMult. Although this method achieves promising performance in this task, there are some limitations. This R-GCN model alone cannot learn relation embeddings, which are very important for knowledge graph applications, since they only define message propagation strategies for node update. On the other hand, despite the fact that R-GCN with an extra decoder can learn relation embeddings for link prediction task, relation information is repeatedly incorporated in both encoder side and decoder side. As a result, the number of parameters increases. In this paper, we concentrated on solving these issues by finding a more reasonable way to extend traditional GCN to KGs.

\section{Conclusion} \label{sec:con}
In this paper we proposed a unified GCN framework (TransGCN) to learn embeddings of relations and entities simultaneously.To handle the heterogeneous characteristics of knowledge graphs when using traditional GCNs, we came up with a novel way of converting a heterogeneous neighborhood into a homogeneous neighborhood by introducing transformation assumptions, e.g., translation and rotation. Under these assumptions, a relation is treated as a transformation operator transforming a head entity to a tail entity. Translation and rotation assumptions were explored and TransE and RotatE model were wrapped in TransGCN framework, respectively. Any other transformation-based method could work as transformation operations. By doing so, nearby nodes with associated relations were aggregated as messages propagated to the center node as traditional GCNs did, which benefited the entity embedding learning. In addition, we explicitly encoded relations in the same GCN framework so that relation embeddings can be seamlessly encoded with entities at the same time. In this sense, our TransGCN framework can be interpreted as a new (knowledge) graph encoder which produces both entity embeddings and relation embeddings. This encoder can be further incorporated into an encoder-decoder framework for any other tasks. Experimental results on two datasets - ~\fb~ and ~\wn~ showed that our unified TransGCN models, both TransE-GCN  and RotatE-GCN models consistently outperformed the baseline - R-GCN model by noticeably large margins in terms of all metrics, which demonstrated the effectiveness of the conversion idea in dealing with heterogeneous neighbors. Additionally, both models performed better than their base models, i.e., TransE and RotatE, showing the significance of explicitly modeling local structural information in knowledge graph embedding learning.

In this paper, although relations are encoded and learned in our GCN framework, they are updated simply by being passed through a separated linear transformation. In the future, we plan to explore approaches to directly operating convolutions on relations so that the local structure of graphs could also play a role in relation embedding learning. In addition, a weighting mechanism should be studied to measure unequal contributions of neighbors.

\bibliographystyle{ACM-Reference-Format}
\bibliography{sample-base}

\end{document}